\documentclass[11pt]{article}
\usepackage{paclic31}
\usepackage{mathtools}
\usepackage{times}
\usepackage{graphics}
\usepackage{latexsym}
\usepackage{amsmath}
\usepackage{multirow}
\usepackage{url}

\title{The Importance of Automatic Syntactic Features in \\Vietnamese Named Entity Recognition}

\author{Thai-Hoang Pham \\
  R\&D Department \\
  Alt Inc \\
  Hanoi, Vietnam \\
  {\tt phamthaihoang.hn@gmail.com} \\\And
  Phuong Le-Hong \\
  College of Science \\
  Vietnam National University in Hanoi \\
  Hanoi, Vietnam \\
  {\tt phuonglh@vnu.edu.vn} \\}
\date{}

\begin{document}
\maketitle
\begin{abstract}
This paper presents a state-of-the-art system for Vietnamese Named Entity Recognition (NER). By incorporating automatic syntactic features with word embeddings as input for bidirectional Long Short-Term Memory (Bi-LSTM), our system, although simpler than some deep learning architectures, achieves a much better result for Vietnamese NER. The proposed method achieves an overall $F_1$ score of $92.05$\% on the test set of an evaluation campaign, organized in late 2016 by the Vietnamese Language and Speech Processing (VLSP) community. Our named entity recognition system outperforms the best previous systems for Vietnamese NER by a large margin.
\end{abstract}

\section{Introduction}
Named entity recognition (NER) is an essential task in natural language processing that falls under the domain of information extraction. The function of this task is
to identify noun phrases and categorize them into a
predefined class. NER is a crucial pre-processing step used in some NLP applications such as question answering, automatic translation, speech processing, and biomedical science. In two shared tasks, CoNLL
2002\footnote{\url{http://www.cnts.ua.ac.be/conll2002/ner/}} and CoNLL
2003\footnote{\url{http://www.cnts.ua.ac.be/conll2003/ner/}}, language independent NER systems were evaluated for English, German, Spanish, and Dutch. These systems focus on four named entity types namely
person, organization, location, and remaining miscellaneous
entities. 

Lately, an evaluation campaign that systematically compared NER systems for the
Vietnamese language has been launched by the Vietnamese Language and Speech Processing (VLSP)\footnote{\url{http://vlsp.org.vn/}} community. They collect data from electronic newspapers on the web and annotate named entities in this corpus. Similar to the CoNLL 2003 share task, there are four named entity types in VLSP dataset: person (PER), organization (ORG), location (LOC), and miscellaneous entity (MISC). 

In this paper, we present a state-of-the-art NER system for
Vietnamese language that uses automatic syntactic features with word embedding in Bi-LSTM. Our
system outperforms the leading system of the VLSP campaign utilizing a number of syntactic and hand-crafted features, and an end-to-end system described in~\cite{Thai-Hoang:2017} that is a combination of Bi-LSTM, Convolutional Neural Network (CNN), and Conditional Random Field (CRF) about $3\%$. To sum up, the  
overall $F_1$ score of our system is 92.05\% as assessed by the standard test
set of VLSP. The contributions of this work consist of:
\begin{itemize}
\item We demonstrate a deep learning model reaching
  the state-of-the-art performance for Vietnamese NER task. By incorporating automatic syntactic features, our system (Bi-LSTM), although simpler than (Bi-LSTM-CNN-CRF) model described in~\cite{Thai-Hoang:2017}, achieves a much better result on Vietnamese NER dataset. The simple architecture also contributes to the feasibility of our system in practice because it requires less time for inference stage. Our best system utilizes part-of-speech, chunk, and regular expression type features with word embeddings as an input for two-layer Bi-LSTM model, which achieves an $F_1$ score of 92.05\%.
\item We demonstrate the greater importance of syntactic features in
  Vietnamese NER compared to their impact in other languages. Those
  features help improve the $F_1$ score of about $18\%$.
\item We also indicate some network parameters such as network size, dropout are likely to affect the performance of our system.
\item We conduct a thorough empirical study on applying common deep
  learning architectures to Vietnamese NER, including Recurrent Neural Network (RNN),
  unidirectional and bidirectional LSTM. These models are also
  compared to conventional sequence labelling models such as Maximum Entropy Markov models (MEMM).
\item We publicize our NER system for research purpose, which
  is believed to positively contributing to the long-term advancement of
  Vietnamese NER as well as Vietnamese language processing.
\end{itemize}

The remainder of this paper is structured as
follows. Section~\ref{sec:relatedWork} summarizes related work on
NER. Section~\ref{sec:models} describes features and model used in our
system. Section~\ref{sec:experiments} gives experimental results and
discussions. Finally, Section~\ref{sec:conclusion} concludes the paper.
\section{Related Works}
\label{sec:relatedWork}
We categorize two main approaches for NER in a large number of research published in the last two decades. The first
approach is characterized by the use of traditional sequence
labelling models such as CRF, hidden markov
model, support vector machine, maximum entropy that are heavily dependent on hand-crafted features~\cite{Florian:2003,Lin:2009,Durrett:2014,Luo:2015}. These systems made an endeavor to exploit external information instead of the available training data such as gazetteers and unannotated data.

In the last few years, deep neural network approaches have gained in popularity dealing with NER task. With the advance of computational power, there has been more and more research that applied deep learning methods to improve performances of their NLP systems. LSTM and CNN are prevalent models used in these architectures. Firstly,~\cite{Collobert:2011} used a CNN over a sequence of word embeddings with a CRF layer on the top. They nearly achieved state-of-the-art
results on some sequence labelling tasks such as POS tagging,
chunking, albeit did not work for NER. To improve the accuracy for recognizing named entities,~\cite{Huang:2015} used Bi-LSTM with CRF layer for
joint decoding. This model also used hand-crafted features to ameliorate its performance. Recently,~\cite{Chiu:2016} proposed a hybrid model that
combined Bi-LSTM with CNN to learn both character-level and word-level
representations. Instead of using CNN to learn character-level features like~\cite{Chiu:2016},~\cite{Lample:2016} used BI-LSTM to capture both character and word-level features. 

For Vietnamese, VLSP community has organized an
evaluation campaign that follows the rules of CoNLL 2003 shared task to systematically compare NER systems. Participating systems have approached this task by both traditional and deep learning architectures. In particular, the first-rank system of the VLSP campaign which achieved an $F_{1}$ score of $88.78\%$ used MEMM with many hand-crafted features~\cite{Phuong:2016}. Meanwhile,~\cite{Son:2016} adopted deep neural networks
for this task. They used the system provided
by~\cite{Lample:2016}, which consists of two types of LSTM models: Bi-LSTM-CRF and Stack-LSTM. Their best system achieved an $F_{1}$ score of $83.80\%$. More recently,~\cite{Thai-Hoang:2017} used an end-to-end system that is a combination of Bi-LSTM-CNN-CRF for Vietnamese NER. The $F_{1}$ score of this system is $88.59\%$ that is competitive with the accuracy of~\cite{Phuong:2016}.
\section{Methodology}
\label{sec:models}
\subsection{Feature Engineering}
\paragraph{Word Embeddings}
We use a word embedding set trained from 7.3GB of 2 million articles collected through a Vietnamese news portal\footnote{\url{http://www.baomoi.com}} by word2vec\footnote{\url{https://code.google.com/archive/p/word2vec/}} toolkit. Details of this word embedding set are described in~\cite{Thai-Hoang:2017}.

\paragraph{Automatic Syntactic Features}
To ameliorate a performance of our system, we incorporate some syntactic features with word embeddings as input for Bi-LSTM model. These syntactic features are generated automatically by some public tools so the actual input of our system is only raw texts. These additional features consist of part-of-speech (POS) and chunk tags that are available in the dataset, and regular expression types that capture common organization and location names. These regular expressions over tokens described particularly in~\cite{Phuong:2016} are shown to provide helpful features for classifying candidate named entities, as shown in the experiments.

\subsection{Long Short-Term Memory}
\begin{figure}[t]
\centering
\resizebox{0.7\linewidth}{!}{
\includegraphics{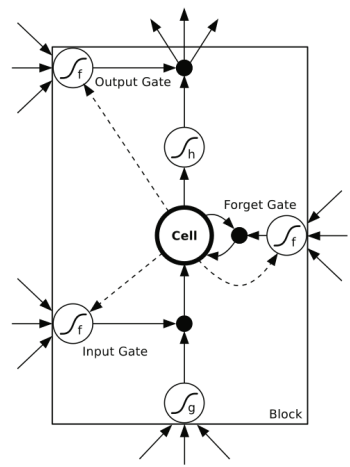}
}
\caption{LSTM memory cell}
\label{fig:2}
\end{figure}

Long short-term memory (LSTM)~\cite{Hochreiter:1997} is a special kind of Recurrent Neural Network (RNN) which is capable of dealing with possible gradient exploding and vanishing problems~\cite{Bengio:1994,Pascanu:2013} when handling long-range sequences. It is because LSTM uses memory cells instead of hidden layers in a standard RNN. In particular, there are three multiplicative gates in a memory cell unit that decides on the amount of information to pass on to the next step. Therefore, LSTM is likely to exploit long-range dependency
data. Each multiplicative gate is computed as follows: 
\begin{align*}
\textbf{i}_{t}&=\sigma(\textbf{W}_{i}\textbf{h}_{t-1}+\textbf{U}_{i}\textbf{x}_{t}+\textbf{b}_{i})\\
\textbf{f}_{t}&=\sigma(\textbf{W}_{f}\textbf{h}_{t-1}+\textbf{U}_{f}\textbf{x}_t+\textbf{b}_{f})\\
\textbf{c}_{t}&=\textbf{f}_{t}\odot \textit{c}_{t-1}+\textbf{i}_{t}\odot\tanh(\textbf{W}_{c}\textbf{h}_{t-1}+\textbf{U}_{c}\textbf{x}_{t}+\textbf{b}_{c})\\
\textbf{o}_{t}&=\sigma(\textbf{W}_{o}\textbf{h}_{t-1}+\textbf{U}_{o}\textbf{x}_{t}+\textbf{b}_{o})\\
\textbf{h}_{t}&=\textbf{o}_{t}\odot\tanh(\textbf{c}_{t})
\end{align*}
where $\sigma$ and $\odot$ are element-wise sigmoid function and
element-wise product, $\textbf{i}$, $\textbf{f}$, $\textbf{o}$ and
$\textbf{c}$ are the input gate, forget gate, output gate and cell
vector respectively. $\textbf{U}_{i}, \textbf{U}_{f}, \textbf{U}_{c},
\textbf{U}_{o}$ are weight matrices that connect input
$\textbf{x}$ and gates, and $\textbf{U}_{i}, \textbf{U}_{f},
\textbf{U}_{c}, \textbf{U}_{o}$ are weight matrices that connect
gates and hidden state $\textbf{h}$, and finally, $\textbf{b}_{i}, \textbf{b}_{f},
\textbf{b}_{c}, \textbf{b}_{o}$ are the bias
vectors. Figure~\ref{fig:2} illustrates a single LSTM memory cell.

\subsection{Bidirectional Long Short-Term Memory}
\begin{figure}[t]
\centering
\resizebox{0.7\linewidth}{!}{
\includegraphics{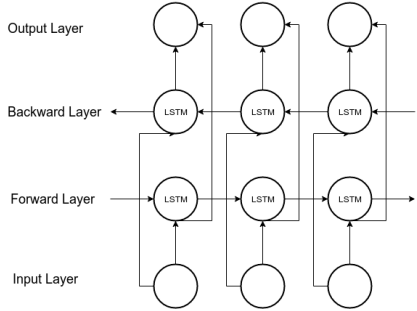}
}
\caption{Bidirectional LSTM}
\label{fig:3}
\end{figure}

The original LSTM uses only past features. For many sequence labelling
tasks, it is beneficial when accessing both past and future
contexts. For this reason, we utilize the bidirectional LSTM
(Bi-LSTM)~\cite{Graves:2005,Graves:2013} for NER task. The basic idea
is running both forward and backward passes to capture past and future
information, respectively, and concatenate two hidden states to form a
final representation. Figure~\ref{fig:3} illustrates the backward and
forward passes of Bi-LSTM.

\subsection{Our Deep Learning Model}
\begin{figure*}[t]
\centering
\includegraphics[scale=0.25]{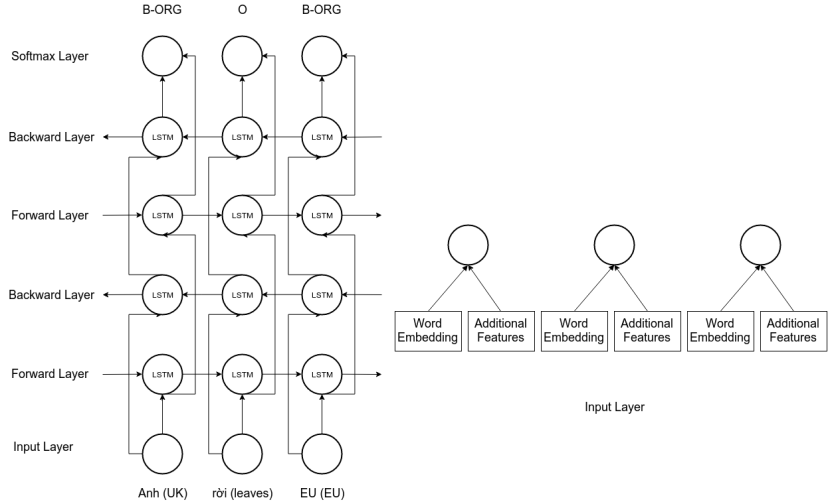}
\caption{Our deep learning model}
\label{fig:4}
\end{figure*}

For Vietnamese named entity recognition, we use a 2-layer Bi-LSTM
with softmax layer on the top to detect named entities in sequence of
sentences. The inputs are the combination of word and syntactic
features, and the outputs are the probability distributions over named
entity tags. Figure~\ref{fig:4} describes the details of our deep learning model. In the next sections, we present our experimental
results. 
\section{Results And Discussions}
\label{sec:experiments}
\subsection{VLSP Corpus}
We conduct experiments on the VSLP NER shared task 2016 corpus. 
Four named entity types are evaluated in this corpus including person, location, organization, and other named entities. Definitions of these entity types match with their descriptions in the CoNLL shared task 2003. 

There are five columns in this dataset including surface word, automatic POS and chunking tags, named entity and nested named entity labels, of which the first four columns conform to the format of the CoNLL 2003 shared task. We do not use the fifth
column because our system focuses
on only named entity without nesting. Named entities are labelled by the IOB notation as in
the CoNLL 2003 shared tasks. In particular, there are 9 named entity labels in this corpus including B-PER and I-PER for persons, B-ORG and I-ORG for organizations, B-LOC and I-LOC for locations, B-MISC and I-MISC for other named entities, and O for other elements. Table~\ref{tab:1} presents the number of annotated named entities in the training and testing set. 
\begin{table}[h]
\center
{
\begin{tabular}{|l|r|r|}
\hline 
Entity Types & Training Set & Testing Set \\ 
\hline 
Location & 6,247 & 1,379 \\ 
\hline 
Organization & 1,213 & 274 \\ 
\hline 
Person & 7,480 & 1,294 \\ 
\hline 
Miscellaneous names & 282 & 49 \\ 
\hline 
All & 15,222 & 2,996 \\ 
\hline 
\end{tabular}}
\caption{Statistics of named entities in VLSP corpus\label{tab:1}}
\end{table}

Because we use early stopping method described in~\cite{Graves:2013} to avoid overfitting when training our neural network models, we hold one part of training data for
validation. The number of sentences of each part of VLSP corpus is described in
Table~\ref{tab:2}.

\begin{table}[h]
\center
\begin{tabular}{|l|r|}
\hline 
Data sets & Number of sentences \\ 
\hline 
Train & 14,861 \\ 
\hline 
Dev & 2,000 \\ 
\hline 
Test & 2,831 \\ 
\hline 
\end{tabular}
\caption{Size of each data set in VLSP corpus}
\label{tab:2} 
\end{table}
\subsection{Evaluation Method}
We evaluate the performance of our system with $F_{1}$ score:
\begin{equation*}
F_{1} = \frac{2*\mathtt{precision}*\mathtt{recall}}{\mathtt{precision}+\mathtt{recall}}
\end{equation*}
Precision and recall are the percentage of correct named entities identified by the system and the percentage of identified named entities present in the corpus respectively. To compare fairly with previous systems, we use an available
evaluation script provided by the CoNLL 2003 shared
task\footnote{\url{http://www.cnts.ua.ac.be/conll2003/ner/}} to calculate $F_{1}$ score of our NER system.
\subsection{Results}
In this section, we analyze the efficiency of word embeddings, bidirectional learning, model configuration, and especially automatic syntactic features.
\paragraph{Embeddings}
To evaluate the effectiveness of word embeddings, we compare the systems
on three types of input: skip-gram, random vector, and one-hot
vector. 

The number of dimensions we choose for word
embedding is 300. We create random vectors for words that do not appear in word embeddings set by
uniformly sampling from the range
$[-\sqrt{\frac{3}{dim}},+\sqrt{\frac{3}{dim}}]$ where $dim$ is the
dimension of embeddings. For random vector setting, we also sample
vectors for all words from this distribution. The performances of the
system with each input type are represented in Table~\ref{tab:3}. 

\begin{table}
\center
\resizebox{\linewidth}{!}{
\begin{tabular}{|l|l|l|l|l|l|l|l|l|l|}
\hline 
Entity & \multicolumn{3}{c|}{Skip-Gram} & \multicolumn{3}{c|}{Random} & \multicolumn{3}{c|}{One-hot} \\ 
\hline 
 & Pre. & Rec. & $F_{1}$ & Pre. & Rec. & $F_{1}$ & Pre. & Rec. & $F_{1}$ \\ 
\hline 
LOC & 83.63 & 82.48 & 83.05 & 79.39 & 66.37 & 72.26 & 79.21 & 72.37 & 75.63 \\ 
MISC & 84.14 & 78.37 & 81.07 & 65.23 & 69.80 & 76.70 & 82.14 & 46.94 & 59.74 \\ 
ORG & 49.85 & 50.51 & 50.07 & 35.19 & 19.56 & 25.11 & 30.56 & 12.04 & 17.28 \\ 
PER & 72.77 & 65.73 & 69.06 & 70.76 & 50.35 & 58.83 & 69.13 & 52.09 & 59.41 \\ 
\hline 
ALL & 75.88 & 72.26 & \textbf{74.02} & 72.99 & 55.23 & 62.87 & 57.68 & 72.88 & 64.39 \\ 
\hline 
\end{tabular}
}
\caption{Performance of our model on three input types}
\label{tab:3}
\end{table}

We can conclude that word embedding is an important factor of our
model. Skip-gram vector significantly improves our performance. The
improvement is about $11\%$ when using skip-gram vectors instead of
random vectors. Thus, we use skip-gram vectors as inputs for our
system.

\paragraph{Effect of Bidirectional Learning}
In the second experiment, we examine the benefit of accessing both
past and future contexts by comparing the performances of RNN, LSTM
and Bi-LSTM models. In this task, RNN model fails because it faces the
gradient vanishing/exploding problem when training with long-range
dependencies (132 time steps), leading to the unstable value of the cost
functions. For this reason, only performances of LSTM and Bi-LSTM
models are shown in Table~\ref{tab:4}.

\begin{table}[h]
\center
\resizebox{\linewidth}{!}{
\begin{tabular}{|l|l|l|l|l|l|l|}
\hline 
Entity & \multicolumn{3}{c|}{Bi-LSTM} & \multicolumn{3}{c|}{LSTM} \\ 
\hline 
 & Pre. & Rec. & $F_{1}$ & Pre. & Rec. & $F_{1}$ \\ 
\hline 
LOC & 83.63 & 82.48 & 83.05 & 74.60 & 77.38 & 75.96 \\ 
MISC & 84.14 & 78.37 & 81.07 & 2.15 & 2.04 & 2.09 \\ 
ORG & 49.85 & 50.51 & 50.07 & 32.22 & 34.60 & 33.60 \\ 
PER & 72.77 & 65.73 & 69.06 & 67.95 & 60.73 & 64.12 \\ 
\hline 
ALL & 75.88 & 72.26 & \textbf{74.02} & 66.61 & 65.04 & 65.80 \\ 
\hline 
\end{tabular}
}
\caption{Performance of our model when using one and two layers}
\label{tab:4}
\end{table}

We see that learning both past and future contexts is very useful for
NER. Performances of all of the entity types are increased, especially
for ORG and MISC. The total accuracy is improved greatly, from
$65.80\%$ to $74.02\%$.

\paragraph{Number of Bi-LSTM Layers}
In the third experiment, we investigate the improvement when
adding more Bi-LSTM layers. Table~\ref{tab:5} shows the accuracy when
using one or two Bi-LSTM layers. We observe a significant improvement when using two layers of
Bi-LSTM. The performance is increased from $71.74\%$ to $74.02\%$

\begin{table}[h]
\center
\resizebox{\linewidth}{!}{
\begin{tabular}{|l|l|l|l|l|l|l|}
\hline 
Entity & \multicolumn{3}{c|}{Two layers} & \multicolumn{3}{c|}{One layer} \\ 
\hline 
 & Pre. & Rec. & $F_{1}$ & Pre. & Rec. & $F_{1}$ \\ 
\hline 
LOC & 83.63 & 82.48 & 83.05 & 82.22 & 80.64 & 81.41 \\ 
MISC & 84.14 & 78.37 & 81.07 & 85.15 & 74.29 & 79.32 \\ 
ORG & 49.85 & 50.51 & 50.07 & 44.10 & 40.88 & 42.39 \\ 
PER & 72.77 & 65.73 & 69.06 & 72.70 & 62.15 & 66.91 \\ 
\hline 
ALL & 75.88 & 72.26 & \textbf{74.02} & 74.83 & 68.91 & 71.74 \\ 
\hline 
\end{tabular}
}
\caption{Performance of our model when using one and two layers}
\label{tab:5}
\end{table}

\paragraph{Effect of Dropout}
In the fourth experiment, we compare the results of our model with and
without dropout layers. The optimal dropout ratio for our experiments is 0.5. 
The accuracy with dropout is $74.02\%$,
compared to $68.27\%$ without dropout. It proves the effectiveness of
dropout for preventing overfitting. 

\begin{table}[h]
\center
\resizebox{\linewidth}{!}{
\begin{tabular}{|l|l|l|l|l|l|l|}
\hline 
Entity & \multicolumn{3}{c|}{Dropout = 0.5} & \multicolumn{3}{c|}{Dropout = 0.0} \\ 
\hline 
 & Pre. & Rec. & $F_{1}$ & Pre. & Rec. & $F_{1}$\\ 
\hline 
LOC & 83.63 & 82.48 & 83.05 & 80.98 & 76.79 & 78.79 \\ 
MISC & 84.14 & 78.37 & 81.07 & 84.09 & 64.49 & 72.73 \\ 
ORG & 49.85 & 50.51 & 50.07 & 41.09 & 32.92 & 36.43 \\ 
PER & 72.77 & 65.73 & 69.06 & 67.35 & 59.23 & 62.97 \\ 
\hline 
ALL & 75.88 & 72.26 & \textbf{74.02} & 71.97 & 64.99 & 68.27 \\ 
\hline 
\end{tabular}
}
\caption{Performance of our model with and without dropout}
\label{tab:6}
\end{table}

\paragraph{Syntactic Features Integration}
As shown in the previous experiments, using only word features in deep learning models is not enough to
achieve the state-of-the-art result. In particular, the accuracy of this model is
only $74.02\%$. This result is far lower in comparison to that of state-of-the-art systems for Vietnamese NER. In the following experiments, we
add more useful features to enhance the performance of our deep learning
model. Table~\ref{tab:7} shows the improvement when adding part-of-speech, chunk,
case-sensitive, and regular expression features. 

\begin{table}[h]
\center
\resizebox{\linewidth}{!}{
\begin{tabular}{|l|l|l|l|}
\hline 
Features & Pre. & Rec. & $F_{1}$ \\ 
\hline 
Word & 75.88 & 72.26 & 74.02 \\ 
\hline 
Word+POS & 84.23 & 87.64 & 85.90 \\ 
\hline 
Word+Chunk & 90.73 & 83.18 & 86.79 \\ 
\hline 
Word+Case & 83.68 & 84.45 & 84.06 \\ 
\hline 
Word+Regex & 76.58 & 71.86 & 74.13 \\ 
\hline 
Word+POS+Chunk+Case+Regex & 90.25 & 92.55 & 91.39 \\ 
\hline 
Word+POS+Chunk+Regex & 91.09 & 93.03 & \textbf{92.05} \\ 
\hline 
\end{tabular}
}
\caption{Performance of our model when adding more features}
\label{tab:7} 
\end{table}

As seen in this table, adding each of these syntactic features
helps improve the performance significantly. The best result we get is adding
part-of-speech, chunk and regular expression features. The accuracy of
this final system is $92.05\%$ that is much higher than $74.02$ of the system without using syntactic features. An explanation for this problem is possibly a characteristic of Vietnamese. In particular, Vietnamese named entities are often a noun phrase chunk.

\paragraph{Comparision with Other Languages}

In the sixth experiment, we want to compare the role of syntactic features for NER task in other languages. For this reason, we run our system on CoNLL 2003 data set for English. The word embedding set we use for English is pre-trained by Glove model and is provided by the authors\footnote{\url{https://nlp.stanford.edu/projects/glove/}}. Table~\ref{tab:9} shows the performances of our system when adding part-of-speech and chunk features.
\begin{table}[h]
\center
\resizebox{\linewidth}{!}{
\begin{tabular}{|l|l|l|l|l|l|l|}
\hline 
• & \multicolumn{3}{c|}{Vietnamese} & \multicolumn{3}{c|}{English} \\ 
\hline 
Features & Pre. & Rec. & $F_{1}$ & Pre. & Rec. & $F_{1}$ \\ 
\hline 
Word & 75.88 & 72.26 & 74.02 & 87.39 & 89.66 & 88.51 \\ 
\hline 
Word + POS + Chunk & 90.39 & 92.59 & 91.48 & 87.08 & 89.59 & 88.31 \\ 
\hline 
\end{tabular} 
}
\caption{The importance of syntactic features for Vietnamese compared to those for English}
\label{tab:9}
\end{table}

For English NER task, adding the syntactic features does not help to improve the performance of our system. Thus, we can conclude that syntactic features have the greater importance in Vietnamese NER compared to their impact in English.
\paragraph{Comparison with Previous Systems}
In VLSP 2016 workshop, there are several different systems proposed for Vietnamese NER. These systems focus on only three
entities types \textit{LOC}, \textit{ORG}, and \textit{PER}. For the purpose of fairness, we evaluate our performances based on these named entity types on the same corpus. The accuracy of our best model over three
entity types is $92.02\%$, which is higher than the best
participating system~\cite{Phuong:2016} in that shared task about $3.2\%$.

Moreover,~\cite{Thai-Hoang:2017} used a combination of Bi-LSTM, CNN, and CRF that achieved the same performance with~\cite{Phuong:2016}. This system is end-to-end architecture that required only word embeddings while~\cite{Phuong:2016} used many syntactic and hand-crafted features with MEMM. Our system surpasses both of these systems by using Bi-LSTM with automatically syntactic features, which takes less time for training and inference than Bi-LSTM-CNN-CRF model and does not depend on many hand-crafted features as MEMM. Table~\ref{tab:8} presents the accuracy of each system.

\begin{table}[h]
\center
\resizebox{0.85\linewidth}{!}{
\begin{tabular}{|l|l|l|}
\hline 
Models & Types & $F_{1}$ \\ 
\hline 
\cite{Phuong:2016} & ME & 88.78 \\ 
\hline 
\cite{Thai-Hoang:2017} & Bi-LSTM-CNN-CRF & 88.59 \\ 
\hline 
Our model & Bi-LSTM & 92.02 \\ 
\hline 
\end{tabular}
}
\caption{Performance of our model and previous systems}
\label{tab:8} 
\end{table}
\section{Conclusion}
\label{sec:conclusion}
In this work, we have presented a state-of-the-art named entity recognition
system for the Vietnamese language, which achieves an $F_1$ score of
$92.05\%$ on the standard dataset published by the VLSP community. Our system outperforms the first-rank
system of the related NER shared task with a large margin, $3.2\%$ in particular. We have also
shown the effectiveness of using automatic syntactic features for Bi-LSTM model that surpass the combination of Bi-LSTM-CNN-CRF models albeit requiring less time for computation.
\section*{Acknowledgement}

The second author is partly funded by the Vietnam National University, Hanoi
(VNU) under project number QG.15.04.  Any opinions, findings and
conclusion expressed in this paper are those of the authors and do not
necessarily reflect the view of VNU.
\bibliographystyle{acl}
\bibliography{bibliography} 
\end{document}